\documentclass[runningheads]{llncs}
\usepackage{graphicx}
\usepackage{amsmath}
\usepackage{amssymb}
\usepackage{algorithm,algpseudocode,float}
\makeatletter
\newenvironment{breakablealgorithm}
  {% \begin{breakablealgorithm}
   \begin{center}
     \refstepcounter{algorithm}% New algorithm
     \hrule height.8pt depth0pt \kern2pt% \@fs@pre for \@fs@ruled
     \renewcommand{\caption}[2][\relax]{% Make a new \caption
       {\raggedright\textbf{\ALG@name~\thealgorithm} ##2\par}%
       \ifx\relax##1\relax % #1 is \relax
         \addcontentsline{loa}{algorithm}{\protect\numberline{\thealgorithm}##2}%
       \else % #1 is not \relax
         \addcontentsline{loa}{algorithm}{\protect\numberline{\thealgorithm}##1}%
       \fi
       \kern2pt\hrule\kern2pt
     }
  }{% \end{breakablealgorithm}
     \kern2pt\hrule\relax% \@fs@post for \@fs@ruled
   \end{center}
  }
\makeatother
\begin{document}
\title{Monte Carlo Neural Fictitious Self-Play: Approach to Approximate Nash Equilibrium of Imperfect-Information Games\thanks{Supported by National Key Research and Development Plan under Grant no. 2016YFB1001203.}}
\titlerunning{MC-NFSP: Approximate Nash Equilibrium of Imperfect Games}
\author{Li Zhang\inst{1} \and
Wei Wang\inst{1} \and
Shijian Li\inst{1,*} \and
Gang Pan\inst{1}}
\authorrunning{L. Zhang et al.}
\institute{Zhejiang University, Hangzhou 310027, China \email{\{zhangli85,21621272,shijianli,gpan\}@zju.edu.cn}}
\maketitle              % typeset the header of the contribution
\begin{abstract}
Researchers on artificial intelligence have achieved human-level intelligence in large-scale perfect-information games, but it is still a challenge to achieve (nearly) optimal results (in other words, an approximate Nash Equilibrium) in large-scale imperfect-information games (i.e. war games, football coach or business strategies). Neural Fictitious Self Play (NFSP) is an effective algorithm for learning approximate Nash equilibrium of imperfect-information games from self-play without prior domain knowledge. However, it relies on Deep Q-Network, which is off-line and is hard to converge in online games with changing opponent strategy, so it can't approach approximate Nash equilibrium in games with large search scale and deep search depth. In this paper, we propose Monte Carlo Neural Fictitious Self Play (MC-NFSP), an algorithm combines Monte Carlo tree search with NFSP, which greatly improves the performance on large-scale zero-sum imperfect-information games. Experimentally, we demonstrate that the proposed Monte Carlo Neural Fictitious Self Play can converge to approximate Nash equilibrium in games with large-scale search depth while the Neural Fictitious Self Play can't. Furthermore, we develop Asynchronous Neural Fictitious Self Play (ANFSP). It use asynchronous and parallel architecture to collect game experience. In experiments, we show that parallel actor-learners have a further accelerated and stabilizing effect on training.

\end{abstract}
\section{Introduction}
With rapid develop of deep reinforcement learning, AI already beats human expert in perfect-information games like Go. However, researchers haven't make same progress in imperfect games like Starcraft or Dota. In order to guarantee effectiveness of our model, we'd better to evaluate training and results in a theorical and quantitive way, but we always neglect it. 

Game theory\cite{Nash.J.:1951} is the cornerstone of human behavior patterns in real world competitions. It studies how agents can maximize their own interests through competition and cooperation, and can measure the quality the decisions in game. It has become an attractive research task in computer science, the intersection research topic called "algorithmic game theory" has established\cite{lavi2007algorithmic}, and gets more and more interact with the development of artificial intelligence\cite{Sanholm.T.Brown.N.:2010,Bosansky}. Its main motivation is to make realworld complext problems, like transaction and traffic control, work in practice.

In Game theory, Nash Equilibrium\cite{Nash.J.:1951} would be an optimal solution in games, i.e. no one can gain extra profit by alleviating their policy. Fictitious play\cite{Brown:1951} is a traditional algorithm for finding Nash Equilibrium in normal-form imperfect games. Fictitious players repeatedly choose best response to the opponent's average strategy. The average strategy of players would converge to Nash Equilibrium. Heinrich et al.\cite{Heinrich.J:2015} proposed Extensive Fictitious Play, extending the idea of fictitious play to extensive-form games. However, the states is represented in the form of look-up table in each tree node, so that the generalization training (of similar states) would be unpractical; And the update of average policy needs the traverse of the whole game tree which results in dimension disaster for large games. Fictitious Self-Play(FSP)\cite{Heinrich.J:2016} addresses these problems by introducing sample–based machine learning approach. The approximation of best response is learned by reinforcement learning and the update of average strategy is processed by sample-based supervised learning. However, due the sampling control, the interaction between agents is controlled by a central controller.

Heinrich and Silver \cite{Heinrich.J:2016} introduced Neural Fictitious Self-Play(NFSP), which combines FSP with neural network function approximation. A player is consisted of Q-learning network and supervised learning network. The algorithm calculates a "best response",by $\epsilon$–greedy deep Q-learning, as well as an average strategy by supervised learning of agents' history behaviors. It  solves the coordinated problem by introducing anticipatory dynamics — players behaves according to a mixture of their average policy and best response.  It's the first end-to-end reinforcement learning method which learns approximate Nash Equilibrium in imperfect games without any prior knowledge.

However, NFSP has bad performance in games with large-scale search space and search depth, because the nature that opponents' strategy is complex and DQN learns in an offline mode. In this paper, we propose Monte Carlo Neural Fictitious Self Play(MC-NFSP). Our algorithm combines NFSP with Monte Carlo Tree Searches\cite{Browne.C.B.:2012}. We evaluate our method in various two-player zero-sum games. Experimentally we show that MC-NFSP would converge to approximate Nash Equilibrium in Othello while NFSP can't.

Another drawback is in NFSP the calculation of best response relies on Deep Q-learning, which takes a long time to run until convergence. In this paper, we propose Asynchronous Neural Fictitious Self-Play(ANFSP), which uses parallel actor learners to stabilize and speed up training. Multiple players choose actions in parallel, on multiple copies of the environment. Players share Q-learning network and supervised learning network, accumulate gradients over multiple steps in Q-learning and calculate gradients of mini-batch in supervised learning. This reduces the data storage memory needed compared to NFSP. We evaluate our method in two-player zero-sum poker games. We show that the ANFSP can approach approximate Nash Equilibrium more stable and quickly compared to NFSP. 

In order to show the effect of the advantage of the techniques of MC-NFSP and ANFSP in more complex game, we also evaluated the effectiveness in a FPS team combat game, in which an AI agent team fights with a human team, and our system provided good tactic strategies and control policies to our agent team, and help it to beat humans. 

\section{Background}

In this section we briefly inroduce: related game theory concepts, current AI systems for games, relationship between reinforcement learning and Nash Equilibrium, and finally the Neural Fictitious Self Play (NFSP) techniques. For a better introduction we refer the reader to\cite{Sutton:1998,Myerson:1991,Heinrich.J:2016} 

\subsection{Related Game Theory Concepts}

\textbf{Game in Study.} In this paper, we mainly research on two-player imperfect-information zero-sum game. A zero-sum game is a game in which the sum of each player's payoff is zero, and an imperfect-information game is a game in which each player only observes partial game state. For example, Texas Hold'em, real-time strategy games and FPS games. Such game is often represented in "Normal form". Normal form is a game representation schema, which lists payoffs that players get as a function of their actions by way of a matrix. In our studied games, players take actions simultaneously. The goal of each player is to maximize their own payoff in the game. Assume $\pi^i(a|U^i)$ is the action distribution of player $i$ given the information set $U^i$ he observes, $\pi = (\pi^1,...,\pi^n)$ refers to the strategy set of all players, ${\Sigma} ^ { i }$ is the behavior set of player $i$, $\pi^{-i}$ is the strategy set in $\pi$ except $\pi^{i}$, $R^i(\pi)$ is the expected payoff the player $i$ gained following strategy $\pi$ in game.  The $\epsilon$-best responses of player $i$ to opponent's strategy $\pi^{-i}$,
\[
BR_{\varepsilon} ^ {i} \left( \pi ^ { - i } \right)=\left\{\pi^ { i } \in \Sigma ^ { i } : R ^ { i } \left( \pi ^ { i } , \pi^{ - i } \right) \geq m a x _ { \pi ^ { \prime } \in \Sigma ^ {i}} R^{ i } \left( \pi ^ { i } , \pi ^ { - i } \right) - \epsilon \right\}
\]
contains all strategies whose payoff against $\pi^{-i}$ that is suboptimal by no more than $\epsilon$.  

\textbf{Nash equilibrium.} Nash equilibrium refers to the strategy that satisfies any player in the game can't obtain higher profit by changing his own strategy when the others don't change their strategy. Nash proved that if we allow mixed strategies, then every game with a finite number of rational players that choose from finitely many pure strategies has at least one Nash equilibrium.

\textbf{Exploitability} evaluates the distance between a strategy and Nash equilibrium strategy, which can be measured from the strategy profits of both side. For two-player zero-sum games, policy $\pi$ is exploitable by $\epsilon$ if and only if
\[
    	\varepsilon = \frac { R ^ { 1 } \left( \mathrm { BR } ^ { 1 } \left( \pi ^ { 2 } \right) , \pi ^ { 2 } \right) + R ^ { 2 } \left( \pi ^ { 1 } , \mathrm { BR } ^ { 2 } \left( \pi ^ { 1 } \right) \right) } { 2 }
\]

In the equation above, $R ^ { 1 } \left( \mathrm { BR } ^ { 1 } \left( \pi ^ { 2 } \right)  \right)$ is the profit(reword) of $player 1$ by making best response to his opponent. It is obvious that an exploitability of $\varepsilon$ yields at least an $\varepsilon$-approximate Nash Equilibrium (distance to Nash Equilibrium no larger than $\varepsilon$), becuase exploitability measures how much the opponent can benefit from a player's failure to adopt a Nash Equilibrium strategy. Nash Equilibrium is unexploitable, i.e. exploitability is 0.

\subsection{Reinforcement learning and Nash Equilibrium}

Reinforcement learning agents learn how to maximize their expected payoff during the interaction with the environment. The interaction can be modelled as Markov Decision Process(MDP). At time step $t$, agent observes current environment state $S_t$ and selects an action  $a_t$ according to policy $\pi$ , where $\pi$ is a mapping from state to action. In return, agent receives reward $r_t$ and next environment state $S_{t+1}$  from environment.The goal of agent is maximizing the accumulated return $G _ { t } = \sum _ { k = 0 } ^ { \infty } \gamma ^ { k } r _ { t + k }$ for each state $S_t$ with discount factor $\gamma\in\left(0,1\right]$. The action-value function $Q^\pi(s,a)=E\left[R_t|s_t=s,a\right]$ defines the expected gain of taking action $a$ in state $s$. It's the common objective function for most reinforcement learning.

An agent is learning on-policy if the policy it learns is what it currently follows, otherwise it's learning off-policy. Deep Q-learning\cite{Volodymyr:2015} is an off-policy method which aims to update the action-value function $Q$ toward the one step return. Monte Carlo Tree Search algorithm\cite{Browne.C.B.:2012} is an on-policy method which aims to choose the best-response action by simulating the game according to policy $\pi$ and updates the action-value function $Q$ till the episode ends. Asynchronous Deep Q-learning\cite{Volodymyr:2016} is a multi-threaded asynchronous variant of Deep Q-learning. It uses multiple actor-learners running in parallel on multiple copies of the environment. Agents share Q-learning network and apply gradient updates asynchronously.

The relationship between reinforcement learning (RL) solution and Game Theory or Nash Equilibrium is: 1) MDP/RL adopts differential learning mechanism, which theoretically achieves Bellman optimality (or Markov perfect equilibrium, a refinement of the concept of Nash equilibrium), so it can learn the subgame optimization substructure including Nash Equilibrium; 2) However, in practice it's very difficult to measure how near a trained strategy to a Nash Equilibrium in large scale games, due to the cost in training.

\subsection{Modern RL systems for games}

In these years, reinforcement learning has great breakthrough in more complex games. The most significant is the DeepMind AlphaGo\cite{silver2016mastering} and AlphaZero\cite{silver2017mastering} which beated world champions in the game Go. AlphaGo is initialized with human anotated training datas, after achieved certain levels, it improves itself by RL and self-play. AlphaZero can teach itself the wining strategy by playing the game purely with itself using a Monte Carlo search tree. DeepMind has shown the effectiveness of Monte Carlo techniques in games, but the game Go is a sequential perfect-information game.

Recently, RL has more researches on imperfect-information games. StarCraft is a hot point of research, many researches like CommNet\cite{sukhbaatar2016learning} and BicNet\cite{peng2017multiagent} focus on small map combat, and DeepMind's AlphaStar\cite{alphastarblog} can play a whole game, and has defeated top human players. The AlphaStar used the supervised training with human data at first, then use a group of agents (league) play with each other in RL to improve to superhuman level, but its current performance is still not stable enough. So it is valuable to think about whether we can get some tools in game theory to measure and control quality of trained strategy, e.g. Nash Equilibrium.

In the study of AI for Poker games, researches often consider Nash Equilibrium. In 2014, Heinrich and Silver of University College London proposed a SmoothUCT algorithm\cite{heinrich2015smooth} that combines the Monte Carlo tree search, converges to approximate Nash equilibrium, and wins three silver medals in the annual Computer Poker Contest (ACPC). In 2015, Lis{\`y} et al. developed an online Counterfactual regret minimization algorithm\cite{lisy2015online}, which can be used to solve Nash Equilibrium in the upper limit betting Texas Hold'em. The artificial intelligence based on this algorithm Cepheus is a near perfect player, human In the long run, the result can only be a tie, or the computer wins. In 2016, Heinrich and Silver proposed the Neural Fictitious Self-Play algorithm, which approximates the Nash equilibrium of imperfect information games without any prior knowledge. 2017 Carnegie Mellon’s artificial intelligence "Libratus"\cite{brown2017libratus} defeated top Texas Hold'em players in one-on-one No-Limit Hold'em, and Libratus developed a balanced game to bring the strategy to Nash equilibrium. Also in 2017, Morav{\v{c}}{\'\i}k et al. proposed DeepStack\cite{moravvcik2017deepstack}, which also defeeted professional human players, and it use exploitability to measure their results.

\subsection{Neural Fictitious Self Play}

Neural Fictitious Self-Play\cite{Heinrich.J:2016} is a model of learning approximate Nash Equilibrium in imperfect-information games. The model combines fictitious play with deep learning. At each iteration, players determine best response to others' average strategy with DQN and update their average strategy by supervised learning. The state-action pair $(S_t, a_t)$ is stored in supervised learning memory only when player determines action from best response. The transition tuple $(S_t,a_t, S_{t+1},r_t)$ is stored in reinforcement learning memory whichever the policy player follows when taking action. NFSP updates the average strategy network with cross entropy loss $L_1$,and updates the best response network with mean squared loss $L_2$.
\[
L_1 \left( \theta ^ { \Pi } \right) = \mathbb { E } _ { ( s , a ) \sim \mathcal { M } _ { S L } } [ - \log \Pi ( s , a | \theta ^ { \Pi } ) ]
\]

\[
L_2  \left( \theta ^ { Q } \right) = \mathbb { E } _ { \left( s , a , r , s ^ { \prime } \right) \sim \mathcal { M } _ { R L } } \left[ \left( r + \max _ { a ^ { \prime } } Q \left( s ^ { \prime } , a ^ { \prime } | \theta ^ { Q ^ { \prime } } \right) - Q ( s , a | \theta ^ { Q } ) \right) ^ { 2 } \right].
\]
 
\begin{figure}[htbp]
\centering
\includegraphics[width=.85\linewidth]{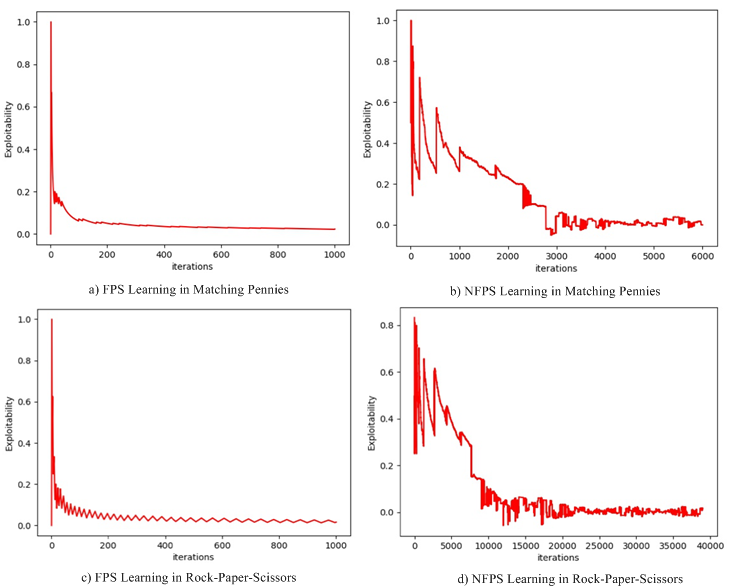}
\caption{Training Efficiency of FP and NFSP}
\label{fig:fig0}
\end{figure}

NFPS use an off-policy methods DQN in it, so it has problems in on-policy games like RTS where we need to sample opponents' changing strategy while we play, and enumerating opponents' strategy is too costly. As shown in Figure \ref{fig:fig0}, we compared the training efficiency of FP and NFSP, subfigures a) and b) show in the game "Matching Pennies", FP converges in 200 iterations, but NFSP in more than 3,000. And in the game "Rock-Paper-Scissors", NFSP converges in more than 10,000 iterations as subfigures c) and d) show.

\section{Monte Carlo Neural Fictitious Self Play}

\subsection{Network Overview}
The Monte Carlo Tree Search (MCTS) algorithm uses a policy network to generate an action probability, and uses a value network to evaluate the value of a state, so it doesn't suffer the complexity in DQN to score each action under each state. So MCTS can be used in high-dimensional and continuous problems. Moreover, MCTS directly uses reward that the player gets after each game to train its networks, it can avoid the inaccurate problem of DQN to evaluate Q-value ($Q(s_{t+1},a|\theta')$) in the early stages. So we combines MCTS and NFSP to propose a new algorithm more suitable to larger imperfect games.

Our aogorithm is called Monte Carlo Neural Fictitious Self Play (MC-NFSP), it learns best response to opponents' average strategy by Monte Carlo Tree Search and updates average strategy by supervised learning with collected best response history. The training dataset is generated from self-play in MC-NFSP. Agent plays a mixture of best response and average strategy as NFSP. Most of time they play an average strategy to the policy $p^{\prime}$ , but with some probability($\eta = 0.1$) they play a best response to MCTS.

\begin{figure}[!]
\centering
\includegraphics[width=.95\linewidth]{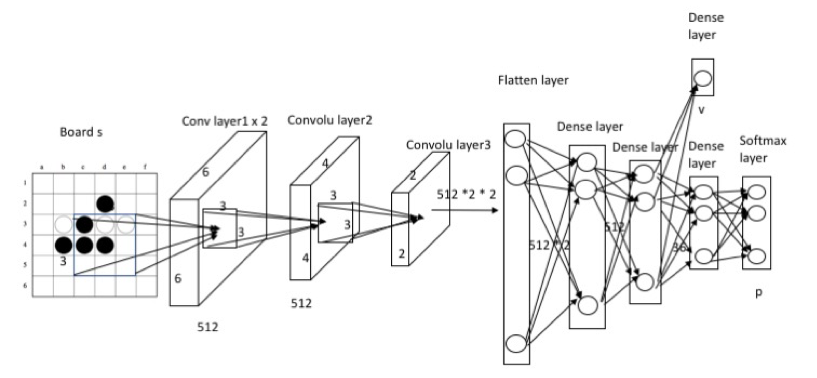}
\caption{Best Response Network for MCTS}
\label{fig:brnet}
\end{figure}

The algorithm makes use of two neural networks: a policy-value network for Monte Carlo Tree Search (i.e. best-response network), a policy network for supervised learning (i.e. average-policy network). The best-response network is shown in Figure. \ref{fig:brnet}. The input is board state. The network has two outputs: a policy $p$, which is a mapping from current state to  action probability, and a value $v$ in $[0,1]$, which is the predicted value of the given state. In our network, relu activation is used in convolution layers; dropout used in fully connection layers to reduce overfitting; and for policy probability, softmax is used. The average-policy network is almost same with best-response ntework except that it only outputs a policy $p^{\prime}$ (no value output), which represent the average policy of player.

\subsection{Algorithm Training}

The networks are trained along with the game self-play. The self-player adopts a mixed policy: $\sigma = ( 1 - \eta ) \Pi + \eta \mathrm { B }$. In each action, the player chooses to use the result from best-response network-based MCTS, or from average-policy network, with a probalilities $\eta, 1- \eta$. 

In MCTS, the algorithm plays a modified Monte Carlo Tree Search(MCTS) algorithm to calculate best response in current state. In MCTS, a tree node is a state, and a edge is an action. MCTS simulates the future plays by adding possible future actionss after an action is executed. At each time, agent chooses action  $a$ maximizing $U(s,a)$.
\[
	U ( s , a ) = Q ( s , a ) + c _ { p u t } \cdot P ( s , a ) \cdot \frac { \sqrt {  N ( s  ) } } { 1 + N ( s , a ) }
\]
$Q(s,a)$ is the expected payoff taking action $a$ in state $s$, $P(s,a)$ is probability of taking action $a$ from state $s$ according to the best-response network. $N(s)$ is the number of visits to state $s$ across simulation. $N(s,a)$ counts the number of times action $a$ been chosen at state $s$ across simulation, $c_{put}$ is a hyperparameter that controls the degree of exploration. Agent takes action $a$ and reaches next state $s^{\prime}$. If $s^{\prime}$ is the terminal state, then the player's final score (win or loss) is used as the node score. If it is not, the opponent takes actions. When a node has not been visited before, when adding it to the game tree, its $P$,$Q$ value is calculated using best-response network as initial node value.

Node value $V$ is propagated along the path visited in current simulation and used to update corresponding $Q(s,a)$ value.
\[
	Q ( s , a ) = ( N ( s , a ) * Q ( s , a ) + V ) / ( N ( s , a ) + 1 )
\]
After multiple simulations, the $N(s,a)$ values are a better approximation for the policy. $\frac { N \left( s _ { \mathrm { b } } \right) } { \sum _ { b } ( N ( s , b ) }$ is normalized as the improved policy $\vec { \pi } ( s )$. Agent picks an action by sampling from the $\vec { \pi } ( s )$. After an episode, tuples $( s , \vec { \pi } ( s ) , v )$ are stored in reinforcement learning memory to train best response network. Pairs $(s, \vec{\pi}(s))$ are stored in supervised learning memory to train average strategy network. After a certain number of episodes, the best response network is trained with loss $l_1$ (in our loss functions, $s_t$ is current game states, $p_t$ is the output of the average network, $z_t$ is the result of the game, value is 1 or -1), the average strategy network is trained with loss $l_2$. 
\[
l_1 = - \sum _ { t } \left( \pi _ { t } \log p _ { t } - \left( \mathrm { v } \left( s _ { t } \right) - z _ { t } \right) ^ { 2 } \right)
\]

\[
l_2 = - \sum _ { t } \vec { \pi } _ { t } \log \vec { p _ { t } }
\]

\begin{breakablealgorithm}
\caption{MC-NFSP algorithm}
\label{alg:MC-NFSP}
\begin{algorithmic}[1]
\State Initialize$\Gamma$,execute function $InitGame()$, $RunAgent(\Pi , B)$;
\Function{InitGame()}{} 
	\State Initialize policy-value network $B(s|\theta^B)$ randomly
	\State Initialize policy network $\Pi ( s | \Theta ^ { \Pi } )$ randomly
	\State Initialize experience replay $M_{RL}$ and $M_{SL}$
	\State (players share networks $B$ and $\Pi$)
\EndFunction
\Function{RunAgent()}{}
	\For{each iteration}
		\State its := its + 1
		\State policy
		$\sigma \leftarrow \left\{
		\begin{aligned}
		B, \quad& with\ probability\ \eta \\
		\pi, \quad & with\ probability\ 1 - \eta
		\end{aligned}
		\right.
		$
		\State observe initial state $s$ and reward $r$
		\While{not terminal}:
			\State If policy comes from $\pi$,choose action $a$ in state $s_t$ according to $\pi$
			\State If policy comes from $B$, choose action according to adapted MCTS 					\State Execute action $a$, observe next state $s_{t+1}$
			\If {$terminal$}:
				\State store $\left( s _ { t } , \vec { \pi _ { t } } , Z _ { t } \right)$ in  $M_{RL}$, store $\left( s _ { t } , \vec { \pi _ { t } } \right)$ in $M_{SL}$ if policy comes from $B$
			\EndIf
		\EndWhile
		\If{$its \% update == 0$}:
			\State update best response network with $l = - \sum _ { t } \left( \vec { \pi _ { t } } \log p _ { t } - \left( \mathrm { v } \left( s _ { t } \right) - z _ { t } \right) ^ { 2 } \right)$
			\State update average network with $l = - \sum _ { t } \vec { \pi _ { t } } \log \vec { p _ { t } }$
		\EndIf
	\EndFor
\EndFunction
\end{algorithmic}
\end{breakablealgorithm}

\subsection{Experiment}

We compare MC-NFSP with NFSP in Othello. Our experiment investigate the convergence of MC-NFSP to Nash equilibrium in Othello and measure the exploitability of learned strategy as comparative standard.  To reduce the calculation time of exploitability, we choose $4\times4$ Othello board.

\subsubsection{Othello}
The neural network in MC-NFSP takes the $4\times4$ board position $s_t$ as input and passes it through two convolutional layers and a flatten layer. Then, the resultant 120 dimensional vector is passed through many fully connected layers, and output both a vector $p$, representing a probability distribution over moves, and a scalar value $v$, representing the probability of the current player winning in position $s_t$ ,for best response network. The architect of average strategy's neural network is same as best response's neural network except the output. It only outputs a vector $p$ representing probability distribution over moves for average strategy network. We set the sizes of memory to 4M and 400K for $M_{RL}$ and $M_{SL}$ respectively. $M_{RL}$ was updated with a circular buffer containing recent training experiences, $M_{SL}$ was updated with reservoir sampling  \cite{Vitter:1985} to ensure an even distribution of training experiences. The reinforcement learning and supervised learning rate were set to 0.01 and 0.005, and both used Adam optimizer. Players perform gradient update every 100 episodes of play. MC-NFSP's anticipatory parameter was set to $\eta = 0.1$. 

\subsubsection{Comparison MC-NFSP with NFSP}
The neural network in NFSP is the same with MC-NFSP' except the output. The output of best response network in NFSP is a vector $Q$, representing the value of each action in state $s_t$. The output of average strategy network is a vector $p$, representing the probability distribution over moves for average strategy network. We set the size of memory to 400w and 40w for $M_{RL}$ and $M_{SL}$ respectively. The reinforcement learning and supervised learning rate were set 0.01 and 0.005. Each player performed gradient updates of mini-batch size 256 per network for every 256 moves. The target network of best response network was refitted every 300 training. NFSP's anticipatory parameter was set to $\eta = 0.1$. The $\epsilon$-greedy policy's exploration rate started at 0.6 and decayed to 0, proportionally to the inverse square root of the number of iterations.

\begin{figure}[htbp]
\centering
\includegraphics[width=.95\linewidth]{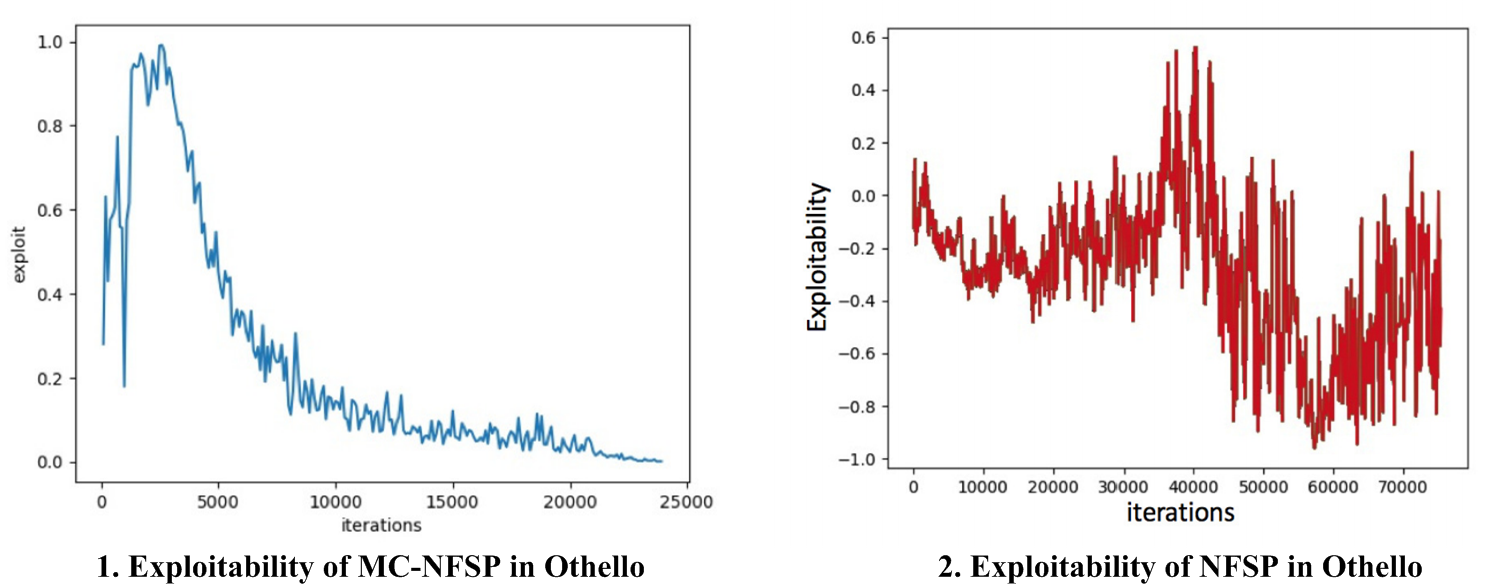}
\caption{Compare MC-NFSP and NFSP in Othello}
\label{fig:fig34}
\end{figure}

Figure \ref{fig:fig34}.2 shows NFSP can't approach Nash equilibrium in Othello. The exploitability of strategy oscillates during the training time. The reason about NFSP doesn't converge to approximate Nash equilibrium in Othello is NFSP players rely on Deep Q-learning approximating best response, which don't have good performance in scenes with large-scale search space and depth.

\section{Asynchronous Neural Fictitious Self Play}
\subsection{Algorithm Overview}

Based on MC-NFSP, we further improve the time-efficiency by proposing a multi-thread learnig mechanism called Asynchronous Neural Fictitious Self-Play(ANFSP), which asynchronously runs multiple players in parallel instances of the game environment. Players run different exploration policies in different threads, and share the nerual network and performs gradient update asynchronously. 

Inspired with A3C algorithm, our algorithm also starts multi-threads of plays, as Algorithm \ref{alg:Asynchronous-Neural-Fictitious} shows. In each thread, an MC-NFSP players a mixture of average-policy and best-response networks. The state-action pair $(S_t, a_t)$ is stored in supervised learning memory only when the player determine action from best response. Each thread computes gradient of best response network using transition tuple $(S_t,a_t, S_{t+1},r_t)$ each step and accumulate gradients over multiple timesteps to certain number before they are applied, which is similar to mini-batches. And we compute gradient of average strategy network using mini-batch of supervised learning memory after multiple timesteps after accumulated to certain number. After a global counter achieves certain count, the networks are updated. The loss of best response network is defined as $l_1$, and the loss of average strategy network is defined as $l_2$.
\[
l_1 = - \left( r + \gamma m a x _ { a ^ { \prime } } Q \left( s ^ { \prime } , a ^ { \prime } ; \theta ^ { Q ^ { - } } \right) - Q \left( s , a ; \theta ^ { Q } \right) \right) ^ { 2 }
\]

\[
l_2 = - \sum _ { i } a _ { i } \log \left( \Pi \left( p _ { i } | s \right) \right)
\]

\begin{breakablealgorithm}
\caption{Asynchronous-Neural-Fictitious-Self-Play}
\label{alg:Asynchronous-Neural-Fictitious}
\begin{algorithmic}[1]
\State InitGame(), Init game $\Gamma$, execute multiple thread $RunAgent()$
\Function{InitGame()}{} 
	\State Init average strategy network $\Pi(s,a|\theta^{\Pi})$
	\State Init Q-value network $Q(s,a|\theta ^ Q)$
	\State Init target network $\theta^{Q^{\prime}} \leftarrow \theta^Q$
	\State Init global anticipatory parameter $\eta$
	\State Init global count T = 0
	\State Init global iteration count iterations = 0
	\State \Return
\EndFunction
\Function{RunAgent()}{}
	\State Init thread count $t \leftarrow 0$
	\Repeat {For each iteration}
		\State policy%
		$\sigma \leftarrow \left\{
		\begin{aligned}
    	\epsilon-greedy(Q),\quad & with\ probability\ \eta \\
		\Pi,\quad & with\ probability\ 1 - \eta
		\end{aligned}
		\right.
		$
		\State observe state $s$ and reward $r$
		\State determine action $a$, observe reward $r_{t+1}$, next state $s_{t+1}$
		\State %
		$y = \left\{ 
		\begin{aligned}
		r \\ 
		r + \gamma m a x _ { a ^ { \prime } } Q \left( s ^ { \prime } , a ^ { \prime } ; \theta ^ { - } \right),\quad & if\ s_{t+1}\ is\ not\ terminal 
		\end{aligned}
		\right. 
		$
		\State accumulate gradient
		$d \theta ^ { Q } \leftarrow d \theta ^ { Q } + \frac { \partial \left( y - Q \left( s , a ; \theta ^ { Q } \right) \right) ^ { 2 } } { d \theta ^ { Q } }$
		\State If policy $\sigma$ comes from $\epsilon-greedy(Q)$, store pair $(s_t, a_t)$ in % $M_{SL}$
		\State $s_{t} \leftarrow s_{t+1}$
		\State $T \leftarrow T+1$
		\State $t \leftarrow t+1$
		\If{$T\ mod\ I_{target} == 0$}
			\State update target network $\theta^{Q^{\prime}} \leftarrow \theta^{Q}$
		\EndIf
		\If{s is terminal}
			\State iterations += 1
			\If{iterations$\ mod\ I_{\text{Asyncupdate }} = = 0$}
				\State update $\theta^Q$ with $d \theta^Q$ asynchronously
				\State update $\theta\Pi$ with $ L \left( \theta ^ { \Pi } \right) = \mathrm { E } _ { ( s , a ) \sim M _ { S L } } [ - \log \Pi ( s , a | \theta ^ { \Pi } ) ]$
				\State $d \theta ^ { Q } \leftarrow 0 , d \theta ^ { \Pi } \leftarrow 0$
			\EndIf
		\EndIf
	\Until{$T > T _ { \max }$}
	\State \Return
\EndFunction
\end{algorithmic}
\end{breakablealgorithm}

\subsection{Experiment}

\subsubsection{Leduc Hold'em}
We compare ANFSP with NFSP in modified two-player Leduc Hold'em. For simplification, we limit the maximum bet size each round in Leduc Hold'em is 2. In the game, the bet history is represented as a tensor with 4 dimensions, namely {players, round, bet, action taken}. Leduc Hold'em contains 2 rounds. Players usually have three actions to choose from, namely {fold, call, raise}. As the game ends once a player gives up, then the betting history is represented as a $2\times2\times2\times2$ tensor. We flatten the 4-dimensional tensor to a vector of length 16. Leduc Hold'em has a card deck of 6 cards. We represent each rounds’ cards by a k-of-n encoding. E.g. LHE has a card vector of 6 cards and we set public cards to 1, the rest to 0. Concatenating with the cards input, we encode the information state of LHE as a vector of length 22.

We started 4 threads with exploration rate randomly chosen from $[0.4,0.6,0.5,0.7]$. The exploration rate decayed to 0, proportionally to the inverse square root of the number of iterations. We set the sizes of $M_{SL}$ to 200w. We train network every 32 iterations of play. We update Deep Q-learning network with accumulated gradients, and update average strategy network with mini-batch size of 128. The reinforcement learning and supervised learning rate were set 0.01 and 0.005, and both used SGD. The target network of Deep Q-learning was updated every 50000 actions. ANFSP's anticipatory parameter was set to $\eta = 0.1$. 

Figure \ref{fig:fig45}.1 shows ANFSP approaching Nash equilibrium in modified Leduc Hold'em. The exploitability declined continually and appeared to stabilize at around 0.64 after 140w episodes of play. The training costed about 2 hours.

\subsubsection{Comparison with NFSP}
The architecture of neural network in NFSP is the same with ANFSP's. We set the size of memory to 20w and 200w for $M_{RL}$ and $M_{SL}$ respectively. The reinforcement learning and supervised learning rate were set 0.01 and 0.005. Players performed gradient updates of mini-batch size 128 per network for every 128 actions. The target network of best response network was refitted every 300 training. NFSP's anticipatory parameter was set to $\eta = 0.1$. The $\epsilon$-greedy policy' exploration rate started at 0.06 and decayed to 0, proportionally to the inverse square root of the number of iterations.

\begin{figure}[h]
\centering
\includegraphics[width=.95\linewidth]{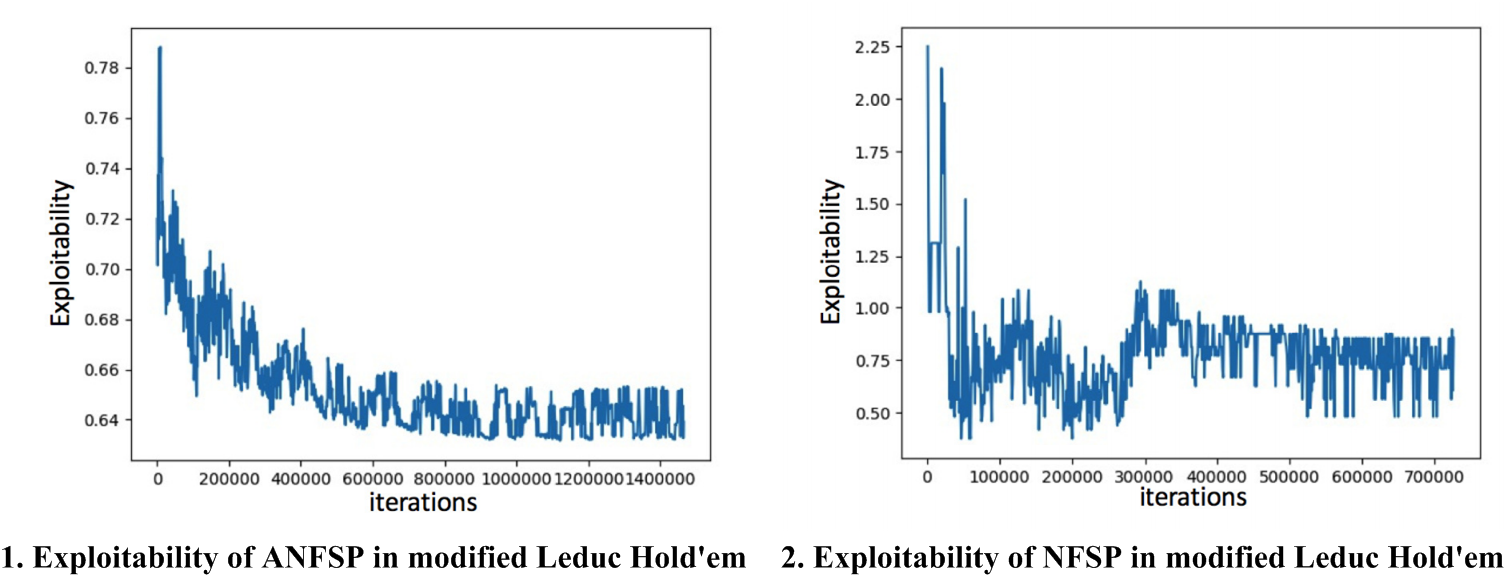}
\caption{Compare ANFSP and NFSP in modified Leduc Hold'em}
\label{fig:fig45}
\end{figure}

Figure \ref{fig:fig45}.2 shows the learning performance of NFSP. The exploitability of strategy fluctuated and appeared to stablize at around 0.75 after 70w episodes of play. The training also costed about 2 hours. It means in same training time (2 hours), ANFSP can complete more episodes and achieve better results (lower exploitability).
\section{Evaluation in First Player Shooting Game}
\subsection{Experiment Setting}
\begin{figure}[!]
\centering
\includegraphics[width=.95\linewidth]{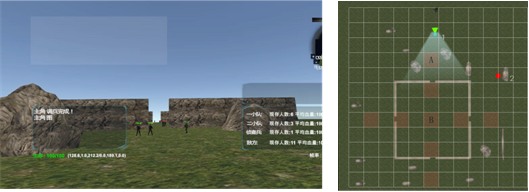}
\caption{FPS Game Environment}
\label{fig:fps}
\end{figure}

In order to evaluate the effectiveness of our algorithm in a complex imperfect-information game, we tried to train it in an FPS game and make it combats with human-bings. The FPS platform used in this experiment is designed by our research team. The game scene is an offensive and defensive confrontation of two teams (10 VS 10). In training, one side is the MC-NFSP, the other side is a memory trained by thousands of human plays (SL-Human). The experiment was performed in a fixed closed 255 x 255 square map. The entire map was divided into 12 x 12 areas each with a 20 x 20 square. The detail of the scene is shown below (Figure. \ref{fig:fps}). All the green areas in Figure. \ref{fig:fps} are passible regions, and the gray areas are obstacles that cannot be crossed (rock or fence). Figure \ref{fig:fps}) is marked with two points A, B, which are the birth points of the two teams. In addition, the ninea reas marked with red represent the destination areas that an agent can choose to affend or defend. The centers of the four walls respectively correspond to four doors. The size of the doors is limited to 2-3 people at the same time. The team outside has a mission to break into the walls and kill all of the inside ones, and the inside team is to defense.

\subsection{Experiment}
\begin{figure}[!]
\centering
\includegraphics[width=.95\linewidth]{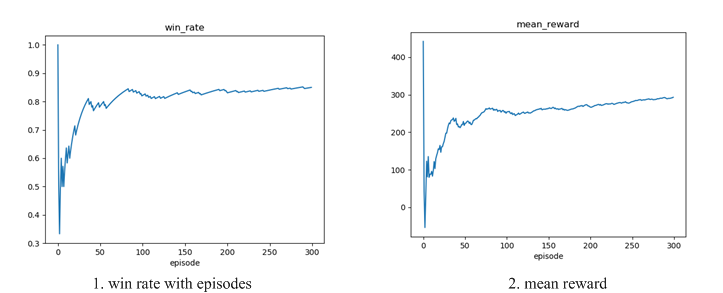}
\caption{Evaluation in FPS Game}
\label{fig:fps_rst}
\end{figure}

In training, each team is represented as a player. The states is a dictionary of the form $\{L,C_{T,L,t},B_{T\prime,L,t}\}$, where $L$ is the location block in game map, $C_{T,L,t}$ means the number of current trained team $T$ in $L$ in time $t$, and $B_{T\prime,L,t}$ is the believed number of team $T\prime$ in location $L$. The actions of a team is the force assignment of number of fighters to different locations like $<n_f, L_1, L_2>$, which means to assign $n_f$ fighters from $L_1$ to $L_2$. For reward, each fighter in team has a health of 100, so the reward of team $T$ is $LostHealth_T-LostHealth_{T\prime}$. Different with our previous works in this paper, the two networks are built and trained for both the outside team and inside team. Figure \ref{fig:fps_rst} shows the training result of the outside team (results of inner team is similar). We can see the training converges very fast (in no more than 150 episode, each episode has 5 games). The win rate of the outside team against the SL-Human gets higher than 80\%, and the loss of training gets near to zero.

After training, we make this algorithm to play the game with pure 10 college students, it plays with one human for 10 rounds (after five rounds, they change the location), totally 100 games are played, in which our trained algorithm achieved 75 victories, so it is a superhuman result.   

\section{Conclusion}
In this paper, we extend the original NFSP algorithm that can learn approximate Nash equilibrium in two-player imperfect-information games, and propose our MC-NFSP and ANFSP algorithms. MC-NFSP use MCTS to help NFSP get rid of offline DQN, so it achieves high improvement on training efficiency, and has approached one more step to achieve approximate Nash equilibrium in larger-scaled imperfect games with wider and deeper game tree. Experiment on Othello shows players' strategy converges to approximate Nash equilibrium with the improved algorithm running a certain round, where the original algorithm cannot converge. ANFSP uses asynchronous and parallel architecture to collect game experiences efficiently and reduces converging time and memory cost that NFSP needed. Experiment on modified Leduc Hold’em shows ANFSP can converge in a shorter time and the convergence is more stable compared with NFSP. Finally we tested the algorithm in FPS games, and it achieved superhuman results in short time, it shows the combination of Monte Carlo Tree Search and NFSP is a practical way to solve imperfect-information game problems.
\bibliographystyle{splncs04}
\bibliography{samplepaper}

\end{document}